\documentclass[a4paper]{article}

\usepackage{INTERSPEECH2019}
\usepackage{amsmath,graphicx,bbm}
\usepackage{multirow, hhline}

\DeclareMathOperator*{\argmin}{arg\,min}

\title{Speaker Adaptation for Attention-Based End-to-End Speech Recognition}
\name{Zhong Meng, Yashesh Gaur, Jinyu Li, Yifan Gong}
\address{Microsoft Corporation, Redmond, WA, USA}
\email{\{zhme,yagaur,jinyli,ygong\}@microsoft.com}

\begin{document}
\maketitle

\begin{abstract}
We propose three regularization-based speaker adaptation
approaches to adapt the attention-based encoder-decoder (AED) model with very
limited adaptation data from target speakers for end-to-end automatic speech recognition. The
first method is Kullback-Leibler divergence (KLD) regularization, in which
the output distribution of a speaker-dependent (SD) AED is forced to be
close to that of the speaker-independent (SI) model by adding a KLD regularization to the
adaptation criterion. To compensate
for the asymmetric deficiency in KLD regularization, an adversarial speaker
adaptation (ASA) method is proposed to regularize the deep-feature distribution of the SD AED through the adversarial learning of an auxiliary discriminator and the SD AED.
The third approach is the multi-task learning, in
which an SD AED is trained to jointly perform the primary task of predicting a large number of output units and an auxiliary task of predicting a small number of output units to
alleviate the target sparsity issue. Evaluated on a Microsoft short message dictation
task, all three methods are highly effective in adapting the AED model,
achieving up to
12.2\% and 3.0\% word error rate improvement over an SI AED trained from
3400 hours data for supervised and unsupervised adaptation, respectively.

\end{abstract}

\section{Introduction}
Recently, remarkable progress has been made in end-to-end (E2E) automatic speech
recognition (ASR) with the advance of deep learning. E2E ASR aims to
directly map a sequence of input speech signal to a sequence of
corresponding output labels as the transcription by incorporating the
acoustic model, pronunciation model and language model in traditional ASR
system into a single deep neural network (DNN). Three dominant approaches to
achieve E2E ASR include: connectionist temporal classification (CTC)
\cite{graves2006connectionist, graves2014towards}, recurrent neural network
transducer \cite{graves2012sequence} and attention-based encoder-decoder
(AED) \cite{chorowski2015attention, bahdanau2016end, chan2016listen}. 

However, the performance of E2E ASR degrades when a
speaker-independent (SI) model is tested with the speech of an unseen
speaker.  A natural solution is to
adapt the SI E2E model to the speech from the target
speaker. The major difficulty for speaker adaptation is that the
speaker-dependent (SD) model  with a large number of
parameters can easily get overfitted to very limited speaker-specific data.

Many methods have been proposed for speaker
adaption in traditional DNN-hidden Markov model
hybrid systems such as
regularization-based \cite{kld_yu, l2_liao, meng2019conditional, multi_huang, toth2016adaptation},
transformation-based \cite{feature_seide, lhuc_pawel_1},
singular value decomposition-based \cite{svd_xue_1,svd_zhao}, 
subspace-based \cite{sc_xue, fhl} and adversarial learning-based \cite{meng2019asa, meng2018speaker} approaches.
Despite the broad success of these methods in hybrid systems, there
has been limited investigation in speaker adaptation for the E2E ASR. In \cite{li2018speaker}, two
regularization-based approaches are shown to be
effective for CTC-based E2E ASR. In \cite{ochiai2018speaker},
constrained re-training \cite{erdogan2016multi} is applied to update a part
of the parameters in a multi-channel AED model. 

In this work, we propose three regularization-based speaker adaptation
approaches for AED-based E2E ASR to overcome the adaptation data
sparsity. We work on the AED model predicting word or subword units (WSUs)
since WSUs have shown to yield better performance than characters as the
output units \cite{gaur2019acoustic, chiu2018state}. The first method is a
Kullback-Leibler divergence (KLD) regularization in which we minimize the KLD between the output distributions of the SD and SI AED models while optimizing the adaptation criterion. 
To offset the deficiency of KLD as an \emph{asymmetric}
distribution-similarity measure \cite{kullback1951information}, we further
propose an adversarial speaker adaptation (ASA) method in which an
auxiliary discriminator network is jointly trained with the SD AED to keep
the deep-feature distribution of the SD AED decoder not far away from that
of the SI AED.
Finally, to address the sparsity of WSU targets in the adaptation data, we propose a
multi-task learning (MTL) speaker adaptation in which an SD
AED is trained to simultaneously perform the primary task of predicting a large number of WSU units and an auxiliary task of predicting a
small number of character units to improve the major task.

We evaluate the three speaker adaptation methods on a Microsoft short
message dictation (SMD) task with 3400 hours live US English training data
and 100-200 adaptation utterances for each speaker.  All three approaches significantly improve over a strong SI AED model. 
In particular, ASA achieves up to 12.2\% and 3.0\% relative word error rate
(WER) gain over the SI baseline for supervised and unsupervised
adaptation, respectively, consistently outperforming the KLD regularization.

\section{Speaker Adaptation for Attention-Based Encoder-Decoder (AED) Model}

We first briefly describe the AED model used
in this work and then elaborate three speaker adaptation methods for AED-based E2E ASR. The SD AED model is always initialized with a
well-trained SI AED predicting WSUs in all three methods.

\subsection{AED Model for E2E ASR}
\label{sec:aed}
In this work, we investigate the speaker adaptation methods for the AED
models \cite{chorowski2015attention, bahdanau2016end, chan2016listen} with
WSUs as the output units. AED model is first introduced in
\cite{cho2014properties, bahdanau2014neural} for neural machine
translation. With the advantage of no conditional independence assumption
over CTC criterion \cite{graves2006connectionist}, AED is introduced, for the first time, to speech area in \cite{chorowski2015attention} for E2E phoneme recognition. In \cite{bahdanau2016end, chan2016listen}, AED is further applied to large vocabulary speech recognition and has recently achieved superior performance to conventional hybrid systems in \cite{chiu2018state}. 

\begin{figure}[htpb!]
	\centering
	\includegraphics[width=0.50\columnwidth]{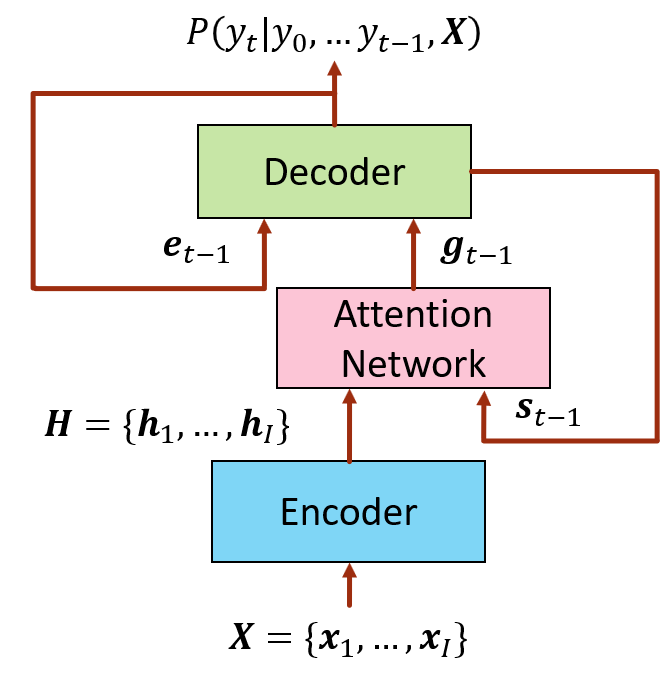}
\vspace{-5pt}
	\caption{The architecture of AED model for E2E ASR.}
	\label{fig:aed}
\vspace{-15pt}
\end{figure}

To achieve E2E ASR, AED directly maps a sequence of speech frames 
 to an output sequence
of WSU labels via an encoder, a decoder and an
attention network as shown in Fig.
\ref{fig:aed}. 

The encoder is an
RNN which encodes the sequence of input speech frames $\mathbf{X}$
into a sequence of high-level features $\mathbf{H} =
\{\mathbf{h}_1, \ldots, \mathbf{h}_T\}$.
AED models the conditional probability distribution
$P(\mathbf{Y} | \mathbf{X})$ over sequences of output WSU labels
$\mathbf{Y}=\{y_1, \ldots, y_T\}$ given a sequence of input speech frames
$\mathbf{X}=\{\mathbf{x}_1, \ldots, \mathbf{x}_I\}$ and, with the encoded
features $\mathbf{H}$, we have
\begin{align}
	P(\mathbf{Y}|\mathbf{X}) = P(\mathbf{Y}|\mathbf{H}) =
	\prod_{t=1}^T P(y_t | \mathbf{Y}_{0:t-1},
	\mathbf{H}) 
\end{align}
A decoder is used to model $P(\mathbf{Y}|\mathbf{H})$.  To capture the
conditional dependence on $\mathbf{H}$, an attention network is used to 
determine
which encoded features in $\mathbf{H}$ should be attended to predict
the output label $y_t$ and to generate a context vector $\mathbf{g}_t$ as a linear combination of $\mathbf{H}$ \cite{chorowski2015attention}.

At each
time step $t$, the decoder RNN takes the sum of the previous WSU embedding 
$\mathbf{e}_{t-1}$ and the context vector $\mathbf{g}_{t-1}$ as the
input to predict the conditional probability of each WSU, i.e., $P(u |
\mathbf{Y}_{0:t-1}, \mathbf{H}), u \in \mathbbm{U}$, at time $t$, where
$\mathbbm{U}$ is the set of all the WSUs:
\begin{align}
        \mathbf{s}_t &= \text{RNN}^{\text{dec}}(\mathbf{s}_{t-1}, \mathbf{e}_{t-1} + 
       \mathbf{g}_{t-1}) \label{eqn:decoder_rnn} \\
        \left[P(u | \mathbf{Y}_{0:t-1}, \mathbf{H})\right]_{u \in
       \mathbbm{U}} &= \text{softmax}\left[W_{y}(\mathbf{s}_t + \mathbf{g}_t) +
       \mathbf{b}_y\right] \label{eqn:decoder_fc}
\end{align}
where $\mathbf{s}_t$ is the hidden state of the decoder RNN. bias $\mathbf{b}_y$ and the matrix $W_y$ are learnable parameters. 

A WSU-based SI AED model is trained to minimize the following loss on the training corpus $\mathbbm{T_r}$.
\begin{align}
	\hspace{-5pt} \mathcal{L}^{\text{WSU}}_{\text{AED}} (\theta^{\text{SI}}, \mathbbm{T_r}) 
	& =-\hspace{-10pt}\sum_{(\mathbf{X}, \mathbf{Y}) \in \mathbbm{T_r}} \sum_{t = 1}^{|\mathbf{Y}|} \log P(y_t |
		\mathbf{Y}_{0:t-1}, \mathbf{H}; \theta^{\text{SI}})
       \label{eqn:aed_loss}
\end{align}
where $\theta^{\text{SI}}$ denotes all the model parameters in the SI AED and $|\mathbf{Y}|$ represents the number of elements in the label sequence
$\mathbf{Y}$.

\subsection{KLD Regularization for Speaker Adaptation}
\label{sec:kld}
Given very limited speech from a target speaker, the SD AED model, usually
with a large number of parameters, can easily get overfitted to the
adaptation data.  To tackle this problem, one solution is to minimize the
KLD between the output distributions of the SI and SD AED models while
training the SD AED with the adaptation data.
We compute the WSU-level KLD between the output distributions of the SI and SD AED models below
\begin{align}
&\hspace{-7pt} \sum_{t=1}^T\sum_{u \in \mathbbm{U}}
P(u|\mathbf{Y}_{0:t-1}, \mathbf{X}; \mathbf{\theta}^{\text{SI}}) \log
\left[ \frac{P(u|\mathbf{Y}_{0:t-1},
	\mathbf{X};\mathbf{\theta}^{\text{SI}})}{P(u|\mathbf{Y}_{0:t-1},
		\mathbf{X}; \mathbf{\theta}^{\text{SD}})}
	\right]
	\label{eqn:kld}
\end{align}
where $\mathbf{\theta}^{\text{SI}}$ denote the all the parameters in SI AED model.

We add only the $\mathbf{\theta}^{\text{SD}}$-related terms to the AED loss as the KLD regularization since $\mathbf{\theta}^{\text{SI}}$ are not updated
during training. Therefore, the
regularized loss function for KLD adaptation of AED is computed below on the adaptation set $\mathbbm{A}$ 
\begin{align}
& \mathcal{L}_{\text{KLD}}(\mathbf{\theta}^{\text{SI}},
\mathbf{\theta}^{\text{SD}}, \mathbbm{A}) = - (1-
\rho)\mathcal{L}^{\text{WSU}}_{\text{AED}}(\mathbf{\theta}^{\text{SD}},
\mathbbm{A}) \nonumber \\
& \qquad \qquad \quad - \rho\sum_{ (\mathbf{X}, \mathbf{Y}) \in \mathbbm{A}} 
        \sum_{t=1}^{|\mathbf{Y}|}\sum_{u \in \mathbbm{U}}
	P(u|\mathbf{Y}_{0:t-1},
	\mathbf{X}; \mathbf{\theta}^{\text{SI}}) \nonumber \\
	& \qquad \qquad \qquad \qquad \qquad \qquad \; \log P(u|\mathbf{Y}_{0:t-1},\mathbf{X};
	\mathbf{\theta}^{\text{SD}}) \nonumber \\
	&=-\sum_{(\mathbf{X}, \mathbf{Y}) \in \mathbbm{A}}
	\sum_{t=1}^{|\mathbf{Y}|}\sum_{u \in \mathbbm{U}} \Bigr\{
	(1-\rho) \mathbbm{1}[u = y_t] \Bigr. \nonumber \\
	& \quad\; \Bigl. + \rho P(u|\mathbf{Y}_{0:t-1},
	\mathbf{X}; \mathbf{\theta}^{\text{SI}}) \Bigr\}
	P(u|\mathbf{Y}_{0:t-1},\mathbf{X};
	\mathbf{\theta}^{\text{SD}}), \\
	& \hat{\mathbf{\theta}}^{\text{SD}} =
	\argmin_{\mathbf{\theta}^{\text{SD}}}
	\mathcal{L}_{\text{KLD}}(\mathbf{\theta}^{\text{SI}},
	\mathbf{\theta}^{\text{SD}}, \mathbbm{A}),
\end{align}
where $\rho \in [0, 1]$ is the regularization weight and
$\mathbbm{1}[\cdot]$ is the indicator function and
$\hat{\mathbf{\theta}}^{\text{SD}}$ denote the optimized parameters. Therefore, KLD regularization for AED is equivalent to using the linear interpolation
between the hard WSU one-hot label and the soft WSU posteriors from SI AED as the new target for standard cross-entropy training.

\subsection{Adversarial Speaker Adaptation (ASA)}
As an \emph{asymmetric} metric, KLD is not a
perfect similarity measure between distributions
\cite{kullback1951information} since the minimization of
$\mathcal{KL}\left(P_{SI} || P_{SD}\right)$ does not guarantee that
$\mathcal{KL}\left(P_{SD} || P_{SI}\right)$ is also minimized. 
Adversarial learning serves as a much better solution since it guarantees that the global optimum is achieved if and only if the SD and SI AEDs share exactly the same hidden-unit distribution  at a certain layer \cite{gan}.
Initially proposed for image generation \cite{gan}, adversarial
learning has recently been widely applied to many aspects of speech area including domain
adaptation \cite{grl_ganin, grl_sun, dsn_meng, meng2018adversarial}, noise-robust ASR \cite{grl_shinohara, grl_serdyuk, meng2018adversarial}, domain-invariant training \cite{meng2018speaker, meng2019aadit, meng2018discriminative}, speech enhancement \cite{pascual2017segan,
meng2018afm, meng2018cycle} and speaker verification \cite{meng2019asv}. ASA is proposed in \cite{meng2019asa} for hybrid system, and in this work, we adapt it to AED-based E2E ASR. 

\begin{figure}[htpb!]
	\centering
	\includegraphics[width=0.95\columnwidth]{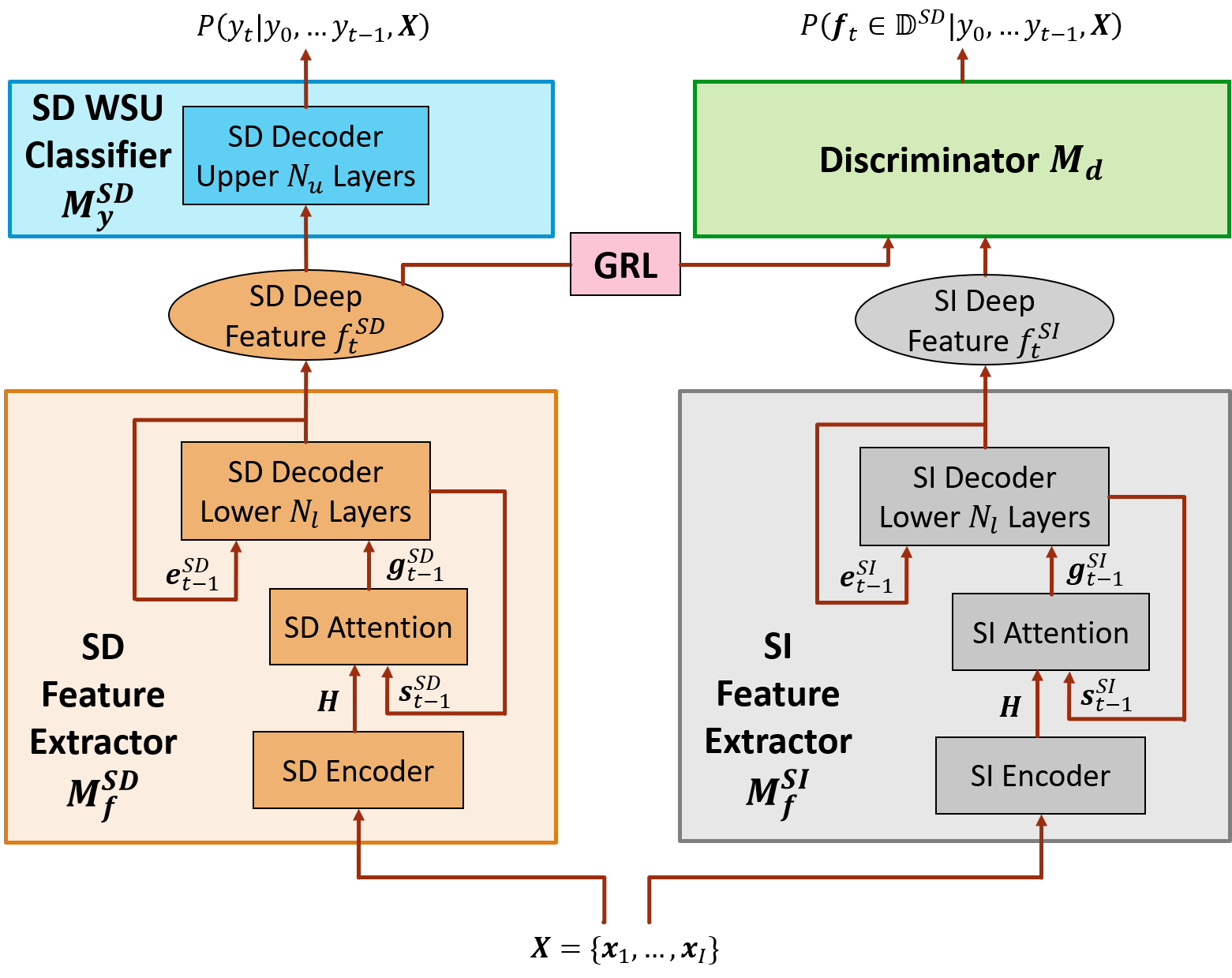}
	\caption{Adversarial speaker adaptation (ASA) of AED model for E2E ASR.
	}
	\label{fig:aed_asa}
	\vspace{-10pt}
\end{figure}

As in Fig.
\ref{fig:aed_asa}, we view the encoder,
the attention network and the first few layers of
the decoder of the SI AED as an SI feature extractor $M_f^{\text{SI}}$ with
parameters $\theta_{f}^{\text{SI}}$ that maps $\mathbf{X}$ to a sequence of
deep hidden features $\mathbf{F}^{\text{SI}} = \{\mathbf{f}_1^{\text{SI}},
\ldots, \mathbf{f}_T^{\text{SI}}\}$ and the rest layers of the SI AED
decoder as a SI WSU classifier $M_y^{\text{SI}}$ with parameters
$\theta_{y}^{\text{SI}}$ (i.e., $\theta^{\text{SI}} =
\{\theta_{f}^{\text{SI}}, \theta_{y}^{\text{SI}}\}$). Similarly, we divide
the SD AED into an SD feature extractor $M_f^{\text{SD}}$ and an SD WSU
classifier $M_y^{\text{SD}}$ in exactly the same way as the SI AED and use
$\theta_{f}^{\text{SI}}$ and $\theta_{y}^{\text{SI}}$ to initialize
$\theta_{f}^{\text{SD}}$ and $\theta_{y}^{\text{SD}}$, respectively (i.e.,
$\theta^{\text{SD}} = \{\theta_{f}^{\text{SD}}, \theta_{y}^{\text{SD}}\}$).
$M_f^{\text{SD}}$ extracts SD deep features $\mathbf{F}^{\text{SD}}$ from
$\mathbf{X}$.

We then introduce an auxiliary discriminator $M_d$ with parameters
$\theta_d$ taking $\mathbf{F}^{\text{SI}}$ and $\mathbf{F}^{\text{SD}}$ as
the input to predict the posterior $P(\mathbf{f}_t \in
\mathbbm{D}^{\text{SD}} | \mathbf{Y}_{0:t-1}, \mathbf{X})$ that
the input deep feature $\mathbf{f}_t$ is generated by the SD AED with the
discrimination loss below.

\begin{align}
& \mathcal{L}_{\text{DISC}}(\theta^{\text{SD}}_f, \theta^{\text{SI}}_f,
\theta_d, \mathbbm{A}) = \nonumber \\
	&        - \sum_{(\mathbf{X}, \mathbf{Y}) \in \mathbbm{A}} \sum_{t
	= 1}^{|\mathbf{Y}|} \left[ \log
	P(\mathbf{f}_t^{\text{SD}} \in \mathbbm{D}^{\text{SD}} |
	\mathbf{Y}_{0:t-1}, \mathbf{X}; \theta^{\text{SD}}_f, \theta_d) \right.\nonumber \\
& \qquad \qquad \qquad \left. + \log
	P(\mathbf{f}_t^{\text{SI}} \in \mathbbm{D}^{\text{SI}} |
\mathbf{Y}_{0:t-1}, \mathbf{X}; \theta^{\text{SI}}_f, \theta_d) \right],
\label{eqn:loss_d}
\end{align}
where $\mathbbm{D}^{\text{SD}}$ and $\mathbbm{D}^{\text{SI}}$ are the
sets of SD and SI deep features, respectively.

With ASA, our goal is to make the distribution of $\mathbf{F}^{\text{SD}}$
similar to that of $\mathbf{F}^{\text{SI}}$ through adversarial training.
Therefore, we minimize $\mathcal{L}_{\text{DISC}}$ with respect to
$\theta_d$ and maximize $\mathcal{L}_{\text{DISC}}$ with respect to
$\theta^{\text{SD}}_f$.  This minimax competition will converge to the
point where $M_f^{\text{SD}}$ generates extremely confusing
$\mathbf{F}^{\text{SD}}$ that $M_d$ is unable to distinguish whether they
are generated by $M_f^{\text{SD}}$ or $M_f^{\text{SI}}$. At the same time, we
minimize the AED loss in Eq. \eqref{eqn:aed_loss} to make
$\mathbf{F}^{\text{SD}}$ WSU-discriminative.  The entire adversarial MTL
procedure of ASA for AED model is formulated below:

\begin{align}
\hspace{-4pt} (\hat{\mathbf{\theta}}_f^{\text{SD}},
	\hat{\mathbf{\theta}}_y^{\text{SD}}) & =
	\argmin_{\mathbf{\theta}_f^{\text{SD}}, \mathbf{\theta}_y^{\text{SD}}}
	\left[\mathcal{L}^{\text{WSU}}_{\text{AED}}(\mathbf{\theta}_f^{\text{SD}},
	\mathbf{\theta}_y^{\text{SD}}, \mathbbm{A}) \right. \nonumber \\
	& \qquad \qquad \qquad \left. - \lambda\mathcal{L}_{\text{DISC}}(\mathbf{\theta}_f^{\text{SD}}, \theta^{\text{SI}}_f,
	\hat{\mathbf{\theta}}_d, \mathbbm{A}) \right] ,
     \label{eqn:minimax_1} \\
     (\hat{\mathbf{\theta}}_d) & =
     \argmin_{\mathbf{\theta}_d}
     \mathcal{L}_{\text{DISC}}(\hat{\mathbf{\theta}}_f^{\text{SD}}, \theta^{\text{SI}}_f,
     \mathbf{\theta}_d, \mathbbm{A}), \label{eqn:minimax_2} 
\end{align}
where $\lambda$ controls the trade-off between
$\mathcal{L}^{\text{WSU}}_{\text{AED}}$ and $\mathcal{L}_{\text{DISC}}$.
Note that the SI AED serves only as a reference network and
$\mathbf{\theta}^{\text{SI}}$ is not updated during training. After
ASA, only the SD AED with adapted parameters $\hat{\theta}^{\text{SD}} =
\{\hat{\theta}_{f}^{\text{SD}}, \hat{\theta}_{y}^{\text{SD}}\}$ are used
for decoding while the auxiliary discriminator $M_d$ is discarded.

\subsection{Multi-Task Learning (MTL) for Speaker Adaptation}
\label{sec:mtl}
One difficulty of adapting AED models is that the WSUs in the adaptation
data are sparsely distributed since the very few adaptation samples are
assigned to a huge number of WSU labels (about 30k). A large proportion
of WSUs are unseen during the adaptation, overfitting the SD AED to a small
space of observed WSU sequences. Inspired by \cite{map_huang,
li2018speaker}, to alleviate this target sparsity issue,
we augment the primary task of predicting a large number of WSU output units
with an auxiliary task of predicting a small number of character output
units (around 30) to improve the primary task via MTL. The adaptation data,
though with a small size, covers a much higher percentage (usually 100\%) of the character
set than that of the WSU set. Predicting the fully-covered character
labels as a secondary task exposes the SD AED to a enlarged acoustic space
and effectively regularizes the major task of WSU prediction. 

We first introduce an auxiliary AED (parameters $\theta^{\text{CHR}}$) with
character output units and initialize its encoder with the encoder
parameters of the WSU-based SI AED $\theta^{\text{SI}}_{\text{enc}}$. Then
we train the decoder (parameters $\theta^{\text{CHR}}_{\text{dec}}$) and
the attention network (parameters $\theta^{\text{CHR}}_{\text{att}}$) of the
character-based AED using all the training data $\mathbbm{T_r}$ to minimize the
character-level AED loss below while keeping its encoder fixed:
\begin{align}
	&\hspace{-1pt} \mathcal{L}^{\text{CHR}}_{\text{AED}}(\theta^{\text{CHR}},
	\mathbbm{T_r}) = 
	-\hspace{-11pt} \sum_{(\mathbf{X}, \mathbf{C}) \in \mathbbm{T_r}} \sum_{l = 1}^{|\mathbf{C}|} P(c_l |
		\mathbf{C}_{0:l-1}, \mathbf{X};
	\theta^{\text{CHR}}) \\
	& (\hat{\theta}^{\text{CHR}}_{\text{dec}}, \hat{\theta}^{\text{CHR}}_{\text{att}}) =
	\argmin_{\theta^{\text{CHR}}_{\text{dec}}, \theta^{\text{CHR}}_{\text{att}}}
	\mathcal{L}^{\text{CHR}}_{\text{AED}}(\theta^{\text{SI}}_{\text{enc}},
		\theta^{\text{CHR}}_{\text{dec}},
	\theta^{\text{CHR}}_{\text{att}}, \mathbbm{T_r}),
       \label{eqn:char_aed_loss}
\end{align}
where $\mathbf{C}= \{c_1, \ldots, c_{L}\}$ is the sequence of character
labels corresponding to $\mathbf{X}$ and $\mathbf{Y}$.

\begin{figure}[htpb!]
	\centering
	\includegraphics[width=0.95\columnwidth]{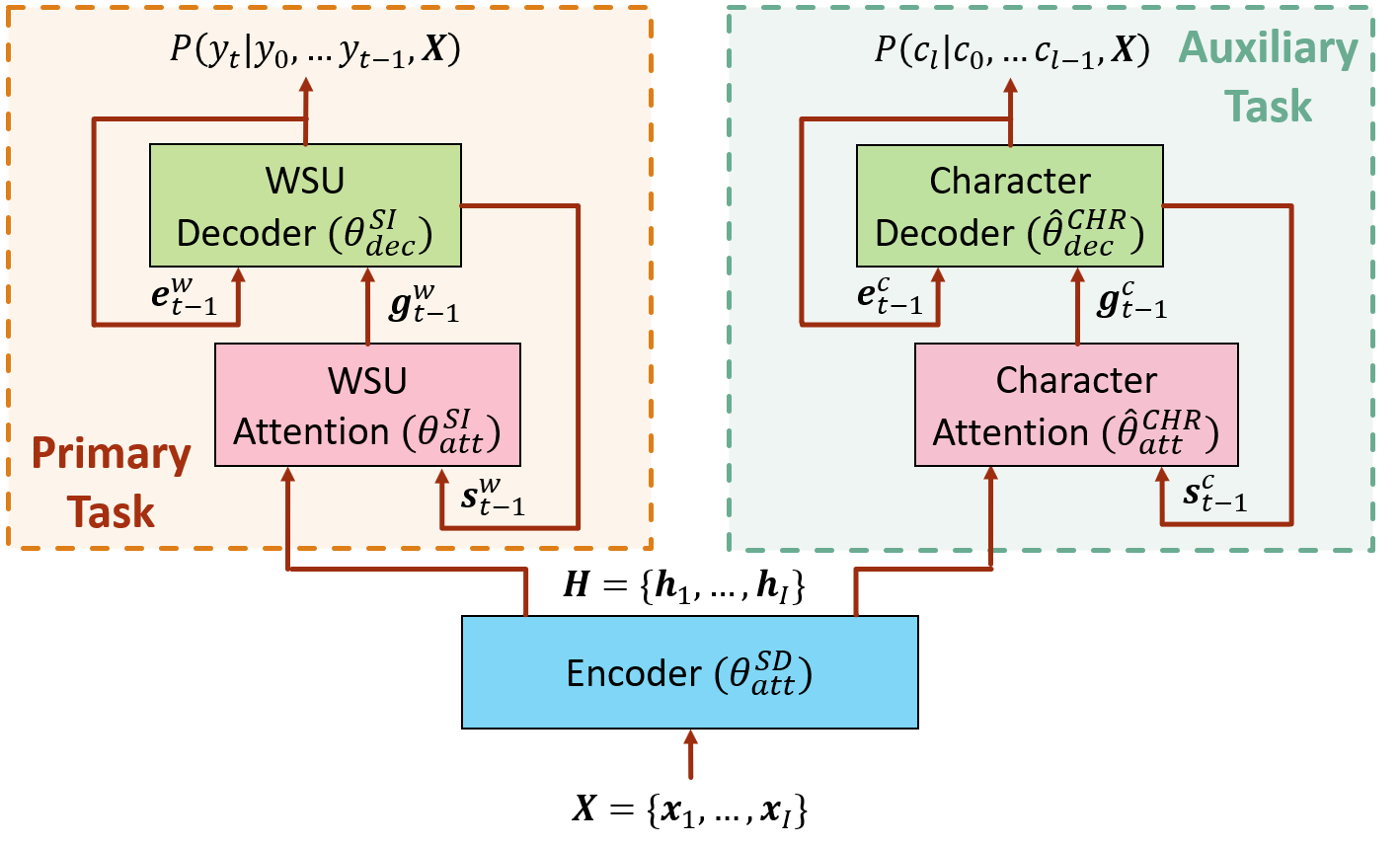}
	\caption{MTL speaker adaptation of AED model for E2E ASR.
	}
	\label{fig:aed_mtl}
\end{figure}

Then we construct an MTL network comprised of the WSU-based SI AED
with initial parameters $\theta^{\text{SI}} =
\{\theta^{\text{SI}}_{\text{enc}}, \theta^{\text{SI}}_{\text{dec}},
\theta^{\text{SI}}_{\text{att}}\}$, a well-trained character-based decoder
with parameters $\hat{\theta}^{\text{CHR}}_{\text{dec}}$ and its attention
network with parameters $\hat{\theta}^{\text{CHR}}_{\text{att}}$ as in Fig.
\ref{fig:aed_mtl}. The
latter two take the encoded features $\mathbf{H}$ from the encoder of the SI AED as the input.  

Finally, we jointly minimize the WSU-level and
character-level AED losses on the adaptation data by updating only the
encoder parameters $\theta^{\text{SD}}_{\text{enc}}$ of the MTL network as
follows:
\begin{align}
	\hat{\theta}^{\text{SD}}_{\text{enc}} &=
	\argmin_{\theta^{\text{SD}}_{\text{enc}}} \left[\beta \mathcal{L}^{\text{WSU}}_{\text{AED}}(\theta^{\text{SD}}_{\text{enc}},
		\theta^{\text{SI}}_{\text{dec}},
\theta^{\text{SI}}_{\text{att}}, \mathbbm{A}) \right. \nonumber \\
& \qquad \qquad \quad \; \left. + (1-\beta) \mathcal{L}^{\text{CHR}}_{\text{AED}}(\theta^{\text{SD}}_{\text{enc}},
	\hat{\theta}^{\text{CHR}}_{\text{dec}},
\hat{\theta}^{\text{CHR}}_{\text{att}}, \mathbbm{A}) \right] 
       \label{eqn:mtl_loss}
\end{align}
where $\beta$ is the interpolation weight for WSU-level AED loss ranging from 0 to 1. 
After the MTL, only the adapted WSU-based SD AED with parameters
$\hat{\theta}^{\text{SD}} = \{\hat{\theta}^{\text{SD}}_{\text{enc}},
\theta^{\text{SI}}_{\text{dec}}, \theta^{\text{SI}}_{\text{att}}\}$ is 
used for decoding. The character-based decoder and attention network are
discarded.

\section{Experiments}
We evaluate the three speaker adaptation methods for AED-based E2E ASR on the Microsoft Windows phone SMD task.

\vspace{-5pt}
\subsection{Data Preparation}
The training data consists of 3400 hours Microsoft internal live US
English Cortana utterances collected via various deployed speech
services including voice search and SMD. The test set consists of 7
speakers with a total number of 20,203 words.  Two adaptation sets of 
100 and 200 utterances per speaker are used for acoustic model
adaptation, respectively. 
We extract 80-dimensional log Mel filter bank (LFB) features from the
speech signal in both the training and test sets every 10 ms over a 25 ms
window.  We stack 3 consecutive frames and stride the stacked
frame by 30 ms to form 240-dimensional input speech frames as in \cite{chiu2018state}.  Following
\cite{li2018advancing}, we first generate 33755 mixed units as the set of
WSUs based on the training transcription and then produce mixed-unit label
sequences as training targets. 

\vspace{-3pt}
\subsection{SI AED Baseline System}
We train a WSU-based AED model as described in Section \ref{sec:aed} for
E2E ASR using 3400 hours training data.  The encoder is a bi-directional
gated recurrent units (GRU)-RNN \cite{cho2014properties,
chung2014empirical} with $6$ hidden layers, each with 512 hidden units.
Layer normalization \cite{ba2016layer} is applied for each hidden layer.
Each WSU label is represented by a 512-dimensional embedding vector. The
decoder is a uni-directional GRU-RNN with 2 hidden layers, each with 512
hidden units, and an output layer predicting posteriors of the 33k WSU. We use GRU instead of long short-term memory (LSTM) \cite{erdogan2016multi, meng2017deep} for RNN because it has less parameters and is trained faster than LSTM with no loss of performance.
We use PyTorch as the tool \cite{paszke2017automatic} for building,
training and evaluating the neural networks. As shown in Table \ref{table:wer}, the baseline SI AED achieves 14.32\%
WER on the test set.

\vspace{-5pt}
\begin{table}[h]
\setlength{\tabcolsep}{4.5 pt}
\centering
\begin{tabular}[c]{c|c|c|c|c||c|c}
	\hline
	\hline
	\multirow{2}{*}{\begin{tabular}{@{}c@{}} System 
		\end{tabular}} & \multirow{2}{*}{\begin{tabular}{@{}c@{}}
		Adapt \\ Param \end{tabular}} & \multirow{2}{*}{\begin{tabular}{@{}c@{}}
		Weight \end{tabular}} & \multicolumn{2}{c||}{Supervised} &
		\multicolumn{2}{c}{Unsupervised} \\
	\hhline{~~~----}	
	& & & 100 & 200 & 100 & 200 \\
	\hline
	SI & - & - & \multicolumn{4}{c}{14.32} \\
	\hline
	\multirow{4}{*}{\begin{tabular}{@{}c@{}} KLD
		\end{tabular}} & \multirow{4}{*}{\begin{tabular}{@{}c@{}}
				All
		\end{tabular}} & $\rho=0.0$ & 14.09 & 13.30 & 14.14 & 14.04\\ 
	\hhline{~~-----}	
	& & $\rho=0.2$ & 13.97 & 13.14 & 14.04 & 14.01 \\
	\hhline{~~-----}	
	& & $\rho=0.5$ & 14.14 & 13.92 & 14.17 & 14.00 \\
	\hhline{~~-----}	
	& & $\rho=0.8$ & 14.31 & 14.23 & 14.81 & 14.14 \\
	\hline
	\multirow{3}{*}{\begin{tabular}{@{}c@{}} ASA
		\end{tabular}} & \multirow{3}{*}{\begin{tabular}{@{}c@{}}
				All
		\end{tabular}} & $\alpha=0.2$ & 13.29 & \textbf{12.58} & 13.99 & 13.92 \\ 
	\hhline{~~-----}	
	& & $\alpha=0.5$ & 13.37 & 12.66 & \textbf{13.95} & \textbf{13.89} \\
	\hhline{~~-----}	
	& & $\alpha=0.8$ & \textbf{13.20} & 12.76 & 13.98 & 13.94 \\
	\hline
	\hline
	\multirow{3}{*}{\begin{tabular}{@{}c@{}} MTL
		\end{tabular}} & \multirow{3}{*}{\begin{tabular}{@{}c@{}}
				Enc 
		\end{tabular}} & $\beta=0.2$ & 13.3 & \textbf{12.71} & 13.93 & 13.87 \\ 
	\hhline{~~-----}	
	& & $\beta=0.5$ & \textbf{13.26} & 12.73 & 13.86 & 13.83 \\
	\hhline{~~-----}	
	& & $\beta=0.8$ & 13.27	& 12.76 & \textbf{13.80} & \textbf{13.77} \\
	\hline
	\hline
\end{tabular}
\vspace{5pt}
\caption{The WERs (\%) of speaker adaptation using KLD, ASA and 
	MTL for AED E2E ASR on Microsoft SMD task with 3400
	hours training data. Each of the 7 test speakers has 100 or 200 adaptation
utterances. In KLD and ASA adaptation, all the parameters of the AED (``All'') are
updated while, in MTL adaptation, only the AED encoder (``Enc'') is updated. }
\vspace{-25pt}
\label{table:wer}
\end{table}


\subsection{KLD Adaptation of AED}
We first perform KLD adaptation of the SI AED with different 
$\rho$ by updating all the parameters in the SD AED. Direct re-training is performed with on regularization
when $\rho=0$.  As shown in Table \ref{table:wer}, for supervised
adaptation, KLD achieves the best WERs, 13.97\% and 13.14\%, at $\rho=0.2$
for both 100 and 200 adaptation utterances with 2.4\% and 8.2\% relative WER
improvements over the SI baseline. The WER increases as $\rho$ continues to
grow. For unsupervised adaptation, KLD achieves the best WERs, 14.04\% $(\rho= 0.2)$ and 14.00\% $(\rho = 0.5)$, for 100 and 200 adaptation utterances,
which improve the SI AED by 2.0\% and 2.2\% relatively. More adaptation
utterances significantly improves the supervised adaptation but only
slightly reduces the WER in unsupervised adaptation since the decoded
one-best path is not as accurate as the forced alignment.

\subsection{Adversarial Speaker Adaptation (ASA) of AED}

To perform ASA of AED, we construct the SI feature extractor
$M_f^{\text{SI}}$ as the first 2 hidden layers of the decoder, the encoder and the
attention network of the SI AED model. The SI senone classifier
$M_y^{\text{SI}}$ is the decoder output layer.  $M_f^{\text{SD}}$ and
$M_y^{\text{SD}}$ are initialized with $M_f^{\text{SI}}$ and
$M_y^{\text{SI}}$. The discriminator $M_d$ is a feedforward DNN with 2
hidden layers and 512 hidden units for each layer.  The output layer of
$M_d$ has 1 unit predicting the posteriors of $\mathbf{f}_t \in
\mathbbm{D}^{\text{SD}}$. $M_f^{\text{SD}}$, $M_y^{\text{SD}}$ and $M_d$
are jointly trained with an adversarial MTL objective as in
Eq. \eqref{eqn:minimax_2}. We update all the parameters in the SD AED.

As shown in Table \ref{table:wer}, for supervised adaptation, ASA achieves
the best WERs, 13.20\% $(\alpha=0.8)$ and 12.58\% $(\alpha=0.2)$,
with 100 and 200 adaptation utterances, which are 7.8\% and
12.2\% relative improvements over the SI AED baseline, respectively. For unsupervised
adaptation, ASA achieves the best WERs, 13.95\% and 13.89\%, both at
$\alpha = 0.5$ with 100 and 200 adaptation utterances, which improves the
SI AED baseline by 2.6\% and 3.0\% relatively. ASA consistently and
significantly outperforms KLD for both supervised and
unsupervised adaptation and for adaptation data of different sizes.
Especially, for supervised adaptation, ASA achieves 5.5\% and 4.3\%
relative improvements over KLD with 100 and 200 adaptation utterances,
respectively.

\subsection{MTL Adaptation of AED}
In MTL, we first train an auxiliary AED with 30 character units as the output using
the training data and then adapt the SI WSU AED by simultaneously
performing WSU and character prediction tasks. The character-based AED share
the same encoder as the WSU-based AED and has a GRU decoder with 2 hidden
layers, each with 512 hidden units.

Table \ref{table:wer} shows that, for supervised adaptation, MTL achieves
best WERs, 13.26\% $(\beta=0.5)$ and 12.71\% $(\beta=0.2)$, with 100 and 200
adaptation utterances, which improves the SI AED baseline by 7.4\% and
11.2\%, respectively. For unsupervised adaptation, MTL achieves best WERs, 
13.80\% and 13.77\%, both at $\beta = 0.8$, which are 3.6\% and 3.8\%
relative improvements over the SI AED baseline, respectively. Note that the
performance of MTL adaptation is not comparable with that of KLD and ASA
since in MTL, only the encoder (consisting of 32.4\% of the whole AED model
parameters) is updated while in KLD and ASA, the whole
AED model is adapted. The KLD and ASA performance can be
remarkably improved by updating only a portion of the entire model parameters.

\section{Conclusion}
In this work, we propose KLD, ASA and MTL approaches for speaker adaptation
in AED-based E2E ASR system. In KLD, we minimize the KLD between the output
distributions of the SD and SI AED models in addition to the AED loss to
avoid overfitting. In ASA, adversarial learning is used to force the deep
features of the SD AED to have similar distribution with those of the SI
AED to offset the asymmetric deficiency of KLD.
In MTL, an additional task of predicting character units is performed in
addition to the primary task of WSU-based AED to resolve the target
sparsity issue.

Evaluated on Microsoft SMD task, all three methods achieve significant
improvements over a strong SI AED baseline for both supervised and
unsupervised adaptation. ASA improves consistently over KLD by updating all
the AED parameters. By adapting only the encoder with 32.4\% of the full
model parameters, the performance of MTL is not comparable
with that of KLD and ASA. Potentially, much larger improvements can be achieved by KLD and ASA by adapting a subset of entire model parameters.

\vfill\pagebreak

\bibliographystyle{IEEEtran}

\bibliography{ref}

\end{document}